\documentclass{article}

\usepackage{arxiv}

\usepackage[utf8]{inputenc} % allow utf-8 input
\usepackage[T1]{fontenc}    % use 8-bit T1 fonts
\usepackage{hyperref}       % hyperlinks
\usepackage{url}            % simple URL typesetting
\usepackage{booktabs}       % professional-quality tables
\usepackage{amsfonts}       % blackboard math symbols
\usepackage{nicefrac}       % compact symbols for 1/2, etc.
\usepackage{microtype}      % microtypography
\usepackage{lipsum}
\usepackage{graphicx}
\graphicspath{ {./images/} }

\title{A Non-equilibrium Thermodynamic Framework Of Consciousness}

\author{
 Natesh Ganesh \\
  ITL, App. \& Comp. Mathematics Division, NIST \\
  Dept. of Physics, University of Colorado, Boulder, CO\\
   \texttt{natesh.ganesh@colorado.edu} \\
  \\
}

\begin{document}
\maketitle
\begin{abstract}
In this paper, we take a brief look at the advantages and disadvantages of dominant frameworks in consciousness studies - functionalist and causal structure theories, and use it to motivate a new non-equilibrium thermodynamic framework of consciousness. The main hypothesis in this paper will be two thermodynamic conditions obtained from the non-equilibrium fluctuation theorems - \textit{TCC 1 and 2}, that the author proposes as necessary conditions that a system will have to satisfy in order to be \emph{conscious}. These descriptions will look to specify the functions achieved by a conscious system and restrict the physical structures that achieve them without presupposing either of the two. These represent an attempt to integrate consciousness into established physical law (without invoking untested novel frameworks in quantum mechanics and/or general relativity). We will also discuss it's implications on a wide range of existing questions, including a stance on the hard problem. The paper will also explore why this framework might offer a serious path forward to understanding consciousness (and perhaps even realizing it in artificial systems) as well as identifying some problems and challenges that lie ahead.
\end{abstract}

% add in platinum standard of consciousness

% keywords can be removed

\section{Introduction}
Consciousness continues to be of one of the most important, interesting and complex question to focus upon. While the study of consciousness has a long and rich history in the field of philosophy, the scientific study of consciousness has become less taboo recently, and made tremendous progress in the field over the last couple of decades, due to significant contributions from disciplines like neuroscience, cognitive science and computer science. Though research interests have continued to grow, fueled by the recent artificial intelligence/machine learning (AI/ML) revolution (reigniting questions around artificial consciousness), the topic of consciousness itself has generally been ignored or dismissed by a majority of those who work in \emph{mainstream} AI as either an unimportant factor for their research goals or accusing work in (artificial) consciousness as distracting flights of fantasy. 

It seems as this trend might change in the near future as leaders in the field of AI recognize the importance of mechanisms of higher level cognition for making progress in AI, their relationship to the 'easy problems' of consciousness and the important work that has been conducted in the field of cognitive science to understand these better (Yoshua Bengio's keynote address at NEURIPS 2019 being an important example of this \cite{Bengio}). While this might not satisfy those who are interested in the phenomenal aspects of our conscious experience, it represents a step forward in the right direction by the larger AI community. In keeping with the (beginnings of a) trend, the author will look to make the case for a non-equilibrium thermodynamic framework of consciousness, it's relationship to the field of AI and the crucial role that computer hardware engineers might have to play in the scientific study of consciousness.

The author would like to take a brief moment (to digress) and explain the journey towards these ideas, hoping that it would elucidate their motivations as an engineer to study and understand the field of consciousness from a more physics based approach. The author's primary research interests lie in the field of artificial intelligence and was lucky enough to be working on his PhD when the field of machine learning became extremely popular. The availability of large amounts of data and cheap compute (in the form of powerful GPU hardware) is one of the driving forces behind this deep learning boom. The demand for compute to train larger and larger models in AI is doubling every 3.5 months (an unprecedented `super-Moore' rate) \cite{OpenAI}, a trend that has shown no sign of slowing down. This comes at a time when Moore's law \cite{Moore}, which powered the first computer revolution has significantly slowed down (and Dennard voltage scaling \cite{Dennard} has completely broken down) as we approach the fundamental physical limits to scaling. This represents a massive problem for state of the art deep learning techniques, some of which that consume a few GWh's of energy (this would be in the ballpark of the amount of energy produced by a nuclear power plant in an hour) to achieve that level of performance on narrow tasks \cite{Strubell}. While these ideas represent massive successes in our quest for building systems that can solve intelligent tasks, they are still a far-cry from the human brain in terms of performance across tasks and energy efficiency. If the human brain remains the holy grail in AI, then we would be remiss to ignore the fact that human brains have consciousness associated with them, dismissing that aspect of it as irrelevant simply because we can compute the functions associated with intelligence, albeit in a inefficent manner. Engineering energy efficient AI hardware is crucial to continue our progress in the field. As we look to build new types of hardware that mimic the human brain, it is necessary to understand the relationship between physical law (that produced human brains), intelligence and consciousness (human or artificial) and the energy efficiency of the hardware realization.

The rest of this paper is organized as follows - In section 2, we will take a very brief look at two very popular frameworks in consciousness - functionalist and causal structure theories, and some of their advantages and drawbacks. In section 3, we explore what might be some necessary conditions for consciousness and propose why thermodynamic descriptions might provide an appropriate language to describe consciousness. In section 4, we introduce non-equilibrium thermodynamic and provide a brief summary of how the fluctuations theorems are related to information processing operations. In section 5, we use the results from the previous section propose two thermodynamic conditions for consciousness TCC1 and TCC2. In section 6, we list a set of important observations about the implications of TCC2. In section 7, we explain the philosophical stance the thermodynamic framework takes towards the hard problem of consciousness. In section 8, we discuss a couple of immediate challenges that are apparent with the framework (and need to be solved) and then move on to the possibility of machine consciousness in section 9, the role of AI hardware in consciousness and finally conclude in section 10 summarizing the ideas in the paper and painting a picture of the path forward. 

\section{Philosophical Frameworks of Consciousness}
There are a number of rich philosophical frameworks of consciousness that ground the popular theories of consciousness like Global Workspace Theory (GWT), Integrated Information Theory (IIT) and Higher Order (HOT) theories. Specifics in the theories themselves are not the focus of this work and a detailed account of GWT, IIT and HOT can be found at \cite{Baar}, \cite{Tononi} and \cite{Carruthers} respectively. We will instead discuss the popular philosophical frameworks that underlie our study of consciousness currently - the \emph{functionalist} picture (as is the case in GWT) and \emph{causal structure} theory (for IIT). 

\paragraph{Functionalism} 
Chalmers defines functional organization as the following \cite{Chalmers0} - \emph{``This is best understood as the abstract pattern of causal interaction between the components of a system, and perhaps between these components and external inputs and outputs. A functional organization is determined by specifying (1) a number of abstract components, (2) for each component, a number of different possible states, and (3) a system of dependency relations, specifying how the states of each component depends on the previous states of all components and on inputs to the system, and how outputs from the system depend on previous component states. Beyond specifying their number and their dependency relations, the nature of the components and the states is left unspecified. A physical system realizes a given functional organization when the system can be divided into an appropriate number of physical components each with the appropriate number of possible states, such that the causal dependency relations between the components of the system, inputs, and outputs precisely reflect the dependency relations given in the specification of the functional organization. A given functional organization can be realized by diverse physical systems. For example, the organization realized by the brain at the neural level might in principle be realized by a silicon system.''} Levin defines functionalism as - \emph{``In the philosophy of mind is the doctrine that what makes something a mental state of a particular type does not depend on its internal constitution, but rather on the way it functions, or the role it plays, in the system of which it is a part.''} \cite{Levin}. According to Putnam's machine state functionalism \cite{Putnam}, \textit{``any creature with a mind can be regarded as a Turing machine (an idealized finite state digital computer), whose operation can be fully specified by a set of instructions (a 'machine table' or program) each having the form: If the machine is in state $S_i$, and receives input $I_j$, it will go into state $S_k$ and produce output $O_l$ (for a finite number of states, inputs and outputs). On either model, however, the mental states of a creature are to be identified with such `machine table states' $(S_1,S_2,…,S_n)$. These states are not mere behavioral dispositions, since they are specified in terms of their relations not only to inputs and outputs, but also to the state of the machine at the time''} \cite{Levin}.

\paragraph{Causal Structure Theory}
On the other hand, in causal structure theories, we have the \emph{''essential element for understanding consciousness is how parts of a system interact. If a system has the `right' kind of causal structure, in other words, if it's elements interact in the `right' way, it is conscious. Otherwise, it is not.''} \cite{Doerig}. Advocates of IIT start from axioms derived from human consciousness and argue that the causal structure of the system (realizing the function) is also important on top of the function \cite{Tononi}, \cite{Tsuchiya} (though it seems like the input-output function being realized, while useful is not central to IIT). Thus advocates of IIT look to use the knowledge of the recurrent structure that underlies human consciousness. IIT distinguishes on the level of consciousness between say a feed-forward and recurrent neural network that might implement the same function (while functionalism does not). IIT predicts the recurrent network  with non-zero $\phi$ to be conscious, and the feedforward network with $\phi=0$ to be non-conscious. As in the case of functionalism, there have been many arguments both for and against this framework \cite{Doerig}, \cite{Tsuchiya}, \cite{Walker}, \cite{Cerullo}, \cite{Aaronson2}. There is also interesting new work in studying contents of experience using category theory in IIT \cite{Saigo}.

%%% Could definitely expand here more if needed.

\section{What do we want from a description of Consciousness?}
A central endeavor in science is to produce descriptions of physical systems that have both explanatory power of existing observations and provide testable predictions that can be used to falsify the framework. Descriptions vary at the level at which they are most useful at and often use different vocabularies. Physics, chemistry, biology, computation, etc can all be viewed as providing descriptions of physical systems at different resolutions. We can view the functionalism and causal structure frameworks similarly as looking to provide descriptions of consciousness in physical systems at different 'levels.' IIT proponents view CST as an improvement on functionalism since they use the extra information of the causal structure in the human brain (which one can safely assume is necessary for human consciousness) on top of the system's input-output relationships to build a better descriptive model \cite{Hoel0}. We can now ask the following question - \textit{what is it that we want from a description of consciousness?}

We will start building towards this description using `some possible necessary conditions for consciousness' prescribed by Scott Aaronson recently \cite{Aaronson} 
\begin{itemize}
    \item [(A1)] Intelligent behavior (passing some sort of Turing test).
    \item [(A2)] Unpredictablity to outside observers, ability to surprise.
    \item [(A3)] ``Not being a giant lookup table or Boltzmann brain.''
    \item [(A4)] ``Full participation in the thermodynamic Arrow of Time'' (Constantly amplifying microscopic degrees of freedom into permanent records).
\end{itemize}

The goal of these conditions for consciousness is to identify systems which are capable of satisfying these conditions and hence considered to be conscious (An approach similar to IIT). The main takeaways from these initial conditions setup by Aaronson would be that both function and the structure of systems matter in the science of consciousness, as well as it's participation in physical law (as prescribed by the arrow of time). IIT in fact does appeal to the importance of structure of the system that realizes particular functions. However they use the phenomenological experience of human consciousness to identify properties of the physical structures that can realize it, to develop their framework and the measure $\phi$  quantify the level (amount) of consciousness in any system. This complexity like $\phi$ measure might end up being too broad and predict a large amount of consciousness in systems like a grid of XOR-gates \cite{Aaronson2}, \cite{Aaronson3}, \cite{Aaronson4} and might be impossible to experimentally verify without pre-supposing the framework to be true \footnote{This can be avoided by stating that IIT is a theory of human consciousness only. It assumes humans are conscious, uses this assumption and the structures in the conscious human brain to define measures like $\phi$ that is correlated to the amount of consciousness in humans. Statements on the consciousness of systems like grid of XOR gates, feedforward and recurrent neural networks should be viewed more as speculation and not as predictions of the framework. In fact, some proponents of this theory like the authors of \cite{Tsuchiya} seem to be defending this weaker version, while others view it as a theory of consciousness in all system leading to confusion in understanding what exactly the claims and testale predictions of IIT are.} \cite{Doerig}. Functionalism too suffers from similar issue since it prescribes mental states based on the functions a system implements, irrespective of the structure that realizes those functions. This would allow for consciousness (to varying degrees) in a very broad range of systems - even though as the result of evolution, the widely accepted conscious entities are biological systems with recurrent neural structures. Kleiner also discusses the issues around both functional and causal structure models, defining consciousness based on static states of a physical systems, rather than as a dynamic trajectory (of states) over time \cite{Kleiner}.

%%% Use Tegmark's table somewhere while disagreeing with some of the principles - note that he talks about rather than asking why is there qualia/why are some arrangements conscious? what arrangement of atoms produce consciousness? - physics line of thinking, the continuing importance of neuroscience.

Another set of possible necessary conditions for consciousness were proposed by Max Tegmark in \cite{Tegmark}. He took a physics point of view towards the problem of consciousness and looked to reduce the question - \emph{why are some arrangements of matter conscious?} (Science is never good with \emph{why} questions) to \emph{what arrangements of matters are conscious?}. These conditions from \cite{Tegmark} were - 
\begin{itemize}
    \item[(T1)] Information principle - A conscious system has substantial information storage capacity.
    \item[(T2)] Dynamics principle - A conscious system has substantial information processing capacity.
    \item[(T3)] Independence principle - A conscious system has substantial independence from the rest of the world.
    \item[(T4)] Integration principle - A conscious system cannot consist of nearly independent parts. 
    \item[(T5)] Autonomy principle - A conscious system has substantial dynamics and independent.
    \item[(T6)] Utility principle - An evolved conscious system records mainly the information useful to it.
\end{itemize}

We can clearly see the overlap between these conditions (T1)-(T6) and (A1)-(A4). All of the above conditions look reasonable except for (T3), which implies that the system dynamics is dominated by forces from within rather than outside the system. The author is not entirely sure of the necessity of such a condition and would rather assume that system dynamics is dependent on forces both inside and outside the system. Tegmark's framework introduced the idea of \emph{perceptronium} - the most general substrate that is subjectively self-aware \cite{Tegmark}. Viewing consciousness as a fundamental state of matter implies that this framework takes a panpsychist position on consciousness (and thus will suffer from the combination problem \cite{Combination}).

In order to move towards an improved description of consciousness that could be extended to systems beyond humans, whose consciousness status is not known beforehand, we have to parse out what such a description would have to achieve (and what IIT strives for). IIT starts from properties of our phenomenological experiences and derives properties of the physical systems that are required to achieve those experiences. These properties are tied to the recurrent causal structures in the human brain that produces consciousness which is measured by the quantity $\phi$ (Of course IIT then extrapolates and assumes consciousness in all systems has to be linked with causal structure without providing sufficient explanation on why the conscious experience of say a large XOR grid should be anything like our own and thus require similar structures). We would thus want a description of consciousness in a physical system to predict both some input-output functions (intelligent behavior) that we expect a conscious system to exhibit and specify the structure(s) a conscious system could employ to realize this behavior, without implicitly assuming either of them in the descriptive conditions. Thus if a framework is set up such that it provides a descriptive condition for consciousness that explains the input-output functions associated with conscious humans and the specific recurrent structure of the human brain that realizes these conditions, then such a framework can be considered as the step forward from the causal structure theories (and consequently functional theories as well). The author will refer to this underlying philosophical position as \textit{constrained functionalism} - where only particular physical realizations of specific functions are hypothesized to be conscious with the constraints imposed by suitable choice of description (rather than all realizations of that function i.e. functionalism). Notice how this picture goes beyond functionalism by restricting certain realizations of a function from being conscious. It does not rule out certain types of structures like causal structure theories do through underlying assumptions based on the biological picture.

The next question would obviously be - \emph{what are the characteristics of the description for such a framework of consciousness?} The characteristics will obviously also imply the level or resolution at which the description operates at. The title of the paper indicates the direction the author wishes to take, but we will work our way towards that answer by teasing out these characteristics that we would want in such a general description of consciousness. Here is a non-exhaustive list that the author will build upon (that are similar to Aaronson's criterion above) 

\begin{itemize}
    \item [(N1)] As discussed above the description should predict for both functions exhibited by platinum-standard conscious systems \cite{Gamez2} and constrain the structures that realize those functions (as in Aaronson's criterion (A1) and (A3)). For example: the description should be able to predict the function and structure with respect to humans, without assuming either.
    \item[(N2)] The descriptions need to strike a fine balance between able to discuss computation in an implementation independent manner (like computational/functional descriptions do) without being overtly broad and placing absolutely no constraints on the type of systems that realize those functions (unlike CST which does place \emph{some but perhaps not enough} restrictions on the causal structure of the implementation) i.e. constrained functionalism. Both functionalism and causal structure frameworks operate at an abstracted level in order to apply broadly across different systems and avoid the challenges faced by psycho-physical identity theories. However this level of abstraction also comes with it's own setbacks.
    \item[(N3)] A good description of consciousness should be tied into established physical law (as in (A4)) and tie into explanations on why accepted conscious systems (like humans, some animals perhaps) have the structure and functions that they do i.e. the evolutionary processes that produced these systems. An important question here would be the need to understand why certain carbon-based organic structures are the material of choice to realize biological consciousness. While a description of consciousness might allow for consciousness to be realized with other material, it has provide this explanation for living systems as well as be able to determine if certain materials simply cannot form the basis of conscious systems under conditions of interest.
    \item[(N4)] A macroscale coarse grained description on which consciousness \emph{emerges}. We use the term emergence in the weaker epistemic sense - where a macroscropic coarse-grained description is emergent if it has greater predictive power than the microscopic fine-grained description. This implicitly also assumes that only systems with a certain degree of complexity (memory) are systems in which we could discuss with respect to consciousnesss (for eg: an electron cannot be conscious since it is not capable of memory) \footnote{Not all complex systems are necessarily conscious. Frameworks that attribute emergence of consciousness to simply increasing complexity tend to be overtly broad and not necessarily the path forward. Some have argued that IIT's $\phi$ measure is correlated to the complexity of the system and thus ends up attributing arbitrarily large amount of consciousness to a sufficiently complex systems like a large XOR grid in a counter-intuitive manner.}. 
\end{itemize}

With the above criterion (N1-N4) in place, the author will propose that \emph{non-equilibrium (NE) thermodynamics} can provide a suitable description of consciousness in physical systems because -
\begin{itemize}
    \item [(a)] NE thermodynamics can provide a macroscale description of a system that is applicable to all physical systems (irrespective of the substrate or implementation). But not all physical systems are capable of satisfying every specified thermodynamic condition.
    \item[(b)] Concepts of information and entropy are already defined and tied to physical quantities such as free-energy and dissipation in thermodynamics (through concepts such as Landauer's Principle). And ofcourse the field of thermodynamics was built to address notions of energy efficiency.
    \item[(c)] It is tied to physical law, with the 2nd law of thermodynamics fundamentally corresponding to the arrow of time. Furthermore the laws of thermodynamics can provide no-go conditions (like perpetual motion machines for example).
    \item[(d)] NE thermodynamics can deal with dynamic trajectories (of states) in open systems over time as opposed to equilibrium thermodynamics which tend to focus on the states of (closed or isolated) systems at equilibrium. This will help address some of the concerns raised in \cite{Kleiner}. Biological systems are ultimately self-organized open systems and are best described by non-equilibrium formulations in thermodynamics.
    \item[(e)] It can be mapped onto a wealth of existing work that has already been performed in the area of neuroscience (predictive processing, free-energy principle) as well as machine learning (Boltzmann and Helmholtz machines).
\end{itemize}

The author will next proceed over the next few sections to introduce the non-equilibrium fluctuation theorems and use it propose some necessary conditions for consciousness in physical systems.

\section{Non-equilibrium Thermodynamics, Fluctuation Theorems, Memory \& Prediction}
Thermodynamics is the branch of physics that deals with heat and temperature, and their relation to energy, entropy, work, radiation, and properties of matter. It was first developed to determine and improve the efficiency of steam engines by Nicolas Carnot in 1824. Since then the field has evolved tremendously and become a cornerstone in modern physics. A detailed introduction to thermodynamics is available in this excellent book by Kondepudi and Prigogine \cite{Kondepudi}. Most of the early work has been focused on systems evolving to thermal equilibrium (which is characterized by minimization of a free-energy term in isolated and closed systems). At equilibrium, the probability of microstate $i$ ($(p(i)$) is given by the Boltzmann distribution

\begin{center}
    $p(i)=\frac{1}{Z(\beta)}e^{-\beta E_i}$
\end{center}

\noindent where $E_i$ is the energy of the $i$-th microstate, $\beta=\frac{1}{k_BT}$ is the inverse temperature for bath temperature $T$ and Boltzmann constant $k_B$. $Z(\beta)$ is the partition function used to normalize the probabilies and given by $Z(\beta)=\displaystyle\sum\limits_j e^{-\beta E_j}$. 

Non-equilibrium thermodynamics which deals with systems not in equilibrium is a much younger field with initial work produced by Lars Onsager \cite{Onsager}, \cite{Onsager2} on reciprocal relations and Ilya Prigogine \cite{Prigogine} on dissipative structures. The field has gone through an incredible revolution in the last few decades as we have moved away from the inequality versions of the 2nd law towards a more general non-equlibrium equality relationships. Some of the most important work in this area has come from Jarzynski \cite{Jarzynski} and Crooks \cite{Crooks}. Unlike equilibrium thermodynamics which focuses on the states of systems, the fluctuation theorems provide relationships between the probability of forward and  reverse trajectories of microstates over a time interval $\tau$ with the heat dissipated into the bath $\Delta Q_{bath}$ at temperature $T$. The relationship can be stated as below

\begin{equation}\label{NEFT}
     \frac{\pi(\gamma; \tau)}{\pi(\gamma^*; \tau) }= \exp \left[ \beta \Delta Q_{bath}(\gamma) \right] 
\end{equation}

\noindent where $\pi(\gamma; \tau)$ is the probability of the forward trajectory (of microstates) $\gamma$ as the system is driven in time $\tau$ (refer to Fig.(1) where $\gamma= i\rightarrow j \rightarrow k$). $\pi(\gamma^*; \tau)$ represents the probability of the reverse trajectory. $\beta$ is the inverse temperature and $\Delta Q_{bath}(\gamma)$ is the heat dissipated into the bath on the forward trajectory. The left-hand side in Eq.(\ref{NEFT}) indicates the irreversibility between the reverse and forward trajectories, and it's relationship to the heat dissipated into the bath. It is thus a generalization of the second law of thermodynamics. It is important to note that this relationship is valid for any transition between two microstates. Note that when the system is in thermal equilibrium, the probability of forward and reverse trajectories of microstates become nearly equal and we can obtain the Boltzmann distribution for the microstates from Eq.(\ref{NEFT}).

\begin{figure}
\begin{center}
\includegraphics[scale=0.75]{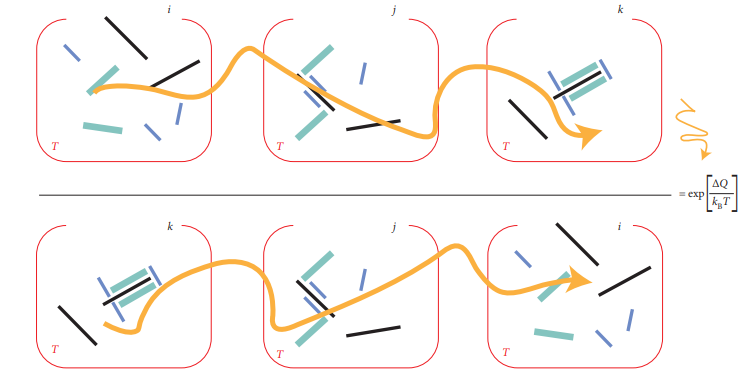}
\end{center}
\caption{\textbf{Dynamical irreversibility and heat production} - The probability of observing a specific sequence of configurations for a system driven with a particular pattern of time-varying field (large arrows) while in contact with a heat reservoir has a set ratio to the probability of observing the time-reversed sequence of events under a time-reversed drive. In particular, these two probabilities differ by an exponential factor of the heat released into the reservoir ($\Delta Q$; small arrow) when the system is driven according to the pattern corresponding to the ratio’s numerator \cite{England2}.}
\end{figure}

There is a lot of ongoing work in adapting these microstate trajectory fluctuation theorems for a wide range of scenarios. Of particular interest to us would be versions of a \emph{macrostate fluctuation theorems} derived in \cite{England1}, \cite{Perunov} and \cite{England2} by integrating over all microstates that constitute a macrostate, and over all relevant trajectories. These can be used to understand the constraints placed on transitions between two arbitrary coarse-grained macroscopic states. The macroscopic fluctuation relation can be stated as 

\begin{equation}\label{MacroFT}
    \frac{\pi(II \rightarrow I; \tau)}{\pi(I \rightarrow II; \tau)}= \langle e^{-\ln \left(\frac{p(i|I)}{p(j|II)} \right)} \langle e^{-\beta \Delta Q_{i\rightarrow j}} \rangle_{i\rightarrow j} \rangle_{I\rightarrow II}
\end{equation}

\noindent where $I$ and $II$ correspond to two macrostate variables defined by underlying conditional microstate distributions $p(i|I)$ and $p(j|II)$ respectively. $\pi(I \rightarrow II; \tau)$ is the probability of the forward trajectory from macrostate $I$ to $II$ in time interval $\tau$. $\Delta Q_{i\rightarrow j}$ is the heat dissipated into the bath during the microstate trajectory $i \rightarrow j$. $\langle \rangle_{i\rightarrow j}$ and $\langle \rangle_{I\rightarrow II}$ corresponds to expectations over all trajectories from microstates $i$ to $j$, and over all $i \in I$ and $j \in II$ respectively. Similar to the above results, a \emph{trajectory-class fluctuation theorems} were proposed in \cite{Wimsatt} and their potential use in thermodynamic computing \cite{Conte} explored. In the next couple of subsections, we will briefly introduce some core ideas that will be necessary for a framework of consciousness and their relationship to the thermodynamic description.

\subsection{Dissipative Adaptation \& Memory}
We will start with the hypothesis of \emph{dissipation driven adaptation}, proposed by the  author in \cite{England2} using versions of Eq.(\ref{MacroFT}), according to which \emph{``..while any given change in shape for the system is mostly random, the most durable and irreversible of these shifts in configuration occur when the system happens to be momentarily better at absorbing and dissipating work. With the passage of time, the `memory' of these less erasable changes accumulates preferentially, and the system increasingly adopts shapes that resemble those in its history where dissipation occurred. Looking backward at the likely history of a product of this non-equilibrium process, the structure will appear to us like it has self-organized into a state that is 'well adapted' to the environmental conditions. This is the phenomenon of dissipative adaptation.''} Thus one can compare the probability of a system in macrostate $I$ evolving to either $II$ or $III$ in some time interval $\tau_0$. These probabilities will depend upon the corresponding amount of work absorbed and dissipated by those transitions, with the more irreversible transition associated with greater dissipation being more likely. This has been studied in greater detail in chemical networks \cite{Horowitz} \cite{Owen}, bistable springs \cite{Hridesh} and spin glasses \cite{Gold}.

The stochastic evolution of the macroscopic variables as prescribed by the fluctuations theorems is an active area of research to better understand the idea of 'memory.' In order to differentiate from retrodiction, Carroll hypothesizes that when macrovariables evolve stochastically, the best a system can do is to use current system state and the past hypothesis to assign probabilities to the past trajectories and thus assign a 'memory' \cite{Carroll1}. Carroll also states in a blog post titled ``Why do we remember the past?'' that \emph{``...memory relies crucially on the second law of thermodynamics. Why do we remember the past and not the future? Because, as entropy increases, we develop correlations between the external universe and our brains; if our universe was in a state of maximum entropy (thermodynamic equilibrium), we wouldn’t be able to remember the past or the future. (We wouldn’t really exist as complex organisms, for that matter; thank the universe for small favors.)''} \cite{Carroll2}. Similar ideas have also been discussed in this excellent post by John Baez in \cite{Baez} bringing together ideas of information, entropy and evolution to view adaptation as `learning the environment.' The major takeaway is that just as we view adaptation is learning the environment over certain spatial and time scales, we can similarly view learning as (dissipation driven) adaptation in the brain over certain spatial and time scales (for eg: plasticity at the level of synapses and neuron populations as adapting to the inputs driving those systems). We can view correlations that are developed between our brains and the external world as our systems evolve as important for the formation of memories. This correlation (and thus the memory) between the current system state $(\mathcal{S})$ and the past environment input history $(\mathcal{H})$ can be quantified using a mutual information measure $\mathcal{I}(\mathcal{S}: \mathcal{H})$. Interestingly the mutual information measure used to capture the amount of memory (here we quantify memory in the sense of capability to store information, and not the colloquial usage of the word which also involves recall), can also be used as a measure of the complexity in the system \cite{Lloyd}, \cite{Crutchfield}, \cite{Shalizi}, \cite{Carroll3}, \cite{Bialek}.

\subsection{Homeostasis \& Prediction}
Another important concept to understand with respect to the macrostate fluctuation relations in Eq.(\ref{MacroFT}) is \emph{homeostasis}. Homeostasis is one of the defining features of biological systems and the process of maintaining a macroscopic variable at a certain (range of) values, exhibiting a natural resistance to outside change. The maintenance of body temperature, pH, blood sugar, potassium levels, etc can all be seen as examples of homeostasis in the human body. Homeostatic mechanisms also play a crucial role in the brain and the nervous system \cite{Turrigiano}. In this subsection, I will obtain the fluctuation theorem version of homeostasis from Eq.(\ref{MacroFT}). 

Let $I$ be the macrostate that is homeostastically maintained over some time interval $\tau_1$ as the system is subject to external driving inputs. Since the macrostate does not change over the time interval, we will substitute $II=I$ in the Eq.(\ref{MacroFT}) from above

\begin{equation}
    \frac{\pi(I \rightarrow I; \tau_1)}{\pi(I \rightarrow I; \tau_1)} = \langle e^{-\ln \left(\frac{p(i|I)}{p(j|I)} \right)} \langle e^{-\beta \Delta Q_{i\rightarrow j}} \rangle_{i\rightarrow j} \rangle_{I\rightarrow I} \nonumber
\end{equation}

\noindent Clearly $\frac{\pi(I \rightarrow I; \tau_1)}{\pi(I \rightarrow I; \tau_1)}=1$, if we are observing the macro-observable $I$ over time, it appears unchanged and there is no difference between a forward $\pi(I \rightarrow I; \tau_1)$ and a reverse $\pi(I \rightarrow I; \tau_1)$. This gives us that

\begin{equation}\label{Homeostasis}
    \langle e^{-\ln \left(\frac{p(i|I)}{p(j|I)} \right)} \langle e^{-\beta \Delta Q_{i\rightarrow j}} \rangle_{i\rightarrow j} \rangle_{I\rightarrow I} = 1
\end{equation}

\noindent Eq.(\ref{Homeostasis}) is a (novel and) general macrostate fluctuation theorem equation for homeostasis. Taking negative logarithm on both side of Eq.(\ref{Homeostasis}) to getd
\begin{equation}\label{Middleqn}
    -\ln \langle e^{-\ln \left(\frac{p(i|I)}{p(j|I)} \right)} \langle e^{-\beta \Delta Q_{i\rightarrow j}} \rangle_{i\rightarrow j} \rangle_{I\rightarrow I} = 0 \nonumber 
\end{equation}

\noindent Following the procedure in \cite{Perunov} and \cite{Ganesh1}, we use the cumulant generating function \footnote{The cumulant generating function is the natural logarithm of the moment generating function i.e. $\ln E[e^{tX}]=\displaystyle\sum\limits_{t=0}^{+\infty} \kappa_n \frac{t^n}{n!}= \mu t+\sigma^2 \frac{t^2}{2}+...$, where $\mu$ and $\sigma^2$ correspond to the mean (1st moment) and variance (2nd moment) respectively.} to write Eq.(\ref{Middleqn}) as

\begin{equation}\label{Main}
    \phi_{I \rightarrow I}=\psi_{I \rightarrow I}
\end{equation}

\noindent where $\phi_{I \rightarrow I}= \frac{1}{\beta} \Delta S_{I \rightarrow I}+ \langle \Delta Q \rangle_{I \rightarrow I}$. $\Delta S_{I \rightarrow I}$ is the change in internal entropy and $\langle \Delta Q \rangle_{I \rightarrow I}$ is the average dissipation into the bath as the system maintains homeostasis. We also have $\psi_{I \rightarrow I}=\phi_{I \rightarrow I} + \frac{1}{\beta}\ln \langle e^{-\ln \left(\frac{p(i|I)}{p(j|I)} \right)} \langle e^{-\beta \Delta Q_{i\rightarrow j}} \rangle_{i\rightarrow j} \rangle_{I\rightarrow I} $, that corresponds to the (sum of the) fluctuations about the mean values. Thus if we imagine a large number of microstates $\{ i\}_{I}$ that all correspond to the observable $I$, then the trajectories of these microstates as they map to other microstates in the same set $\{ i\}_I$ (since $I$ is homeostatically maintained when driven by external inputs) have a dissipation $\Delta Q_{i\rightarrow j}$ associated with that trajectory . We can thus calculate a mean dissipation over all trajectories and microstates $\{i\}_I$, as well the variance and higher order fluctuations about this average.

%%% say some more of what this condition means in normal words

Let's consider a system $\mathcal{S}$ with a certain amount of memory of past history $\mathcal{H}$ quantified as $\mathcal{I}(\mathcal{S}:\mathcal{H})$ (as discussed in the previous subsection). Furthermore let us assume that $\mathcal{S}$ is homeostatic with respect to a macroscopic variable $I$ that is maintains over some time interval $\tau$ and thus satisfies Eq.(\ref{Main}) (and that $\Delta S_{I \rightarrow I}=0$ for now). Consider a very special case when we have $\phi_{I \rightarrow I}=\psi_{I \rightarrow I}$ to both be small (perhaps tending towards some lower bound?). This means that we have both the average dissipation over all trajectories, as well the sum of the fluctuations about this average to be very small (if we assume that variance is the only non-negligible moment in $\phi_{I \rightarrow I}$, then we have that the dissipation associated with all trajectories $i \rightarrow j$ for all $i,j \in \{i\}_I$ is centered around the mean and thus also low). If a system satisfies this condition of low $\phi_{I \rightarrow I}=\psi_{I \rightarrow I}$, then we can show that there exists a coarse graining (what that coarse-graining is, we don't know but it exists) of the microstates of $\mathcal{S}$ into computational states (Fig.(2)), such that the encoding representation of the $i^{\mathcal{H}}$-th input of history $\mathcal{H}$ in the $k^{\mathcal{S}}$-th computational state of the system $\mathcal{S}$ (obtained by coarse-graining over microstates of $I$ using a suitable choice of computationally relevant observable and given by the probability $p(k^{\mathcal{S}}|i^{\mathcal{H}})$) at any time instant in the time interval satisfies the following constraint optimization problem as seen in \cite{Still1}, \cite{Ganesh1}, \cite{Still2}, \cite{Still3} - 

\begin{figure}
\begin{center}
\includegraphics[scale=0.55]{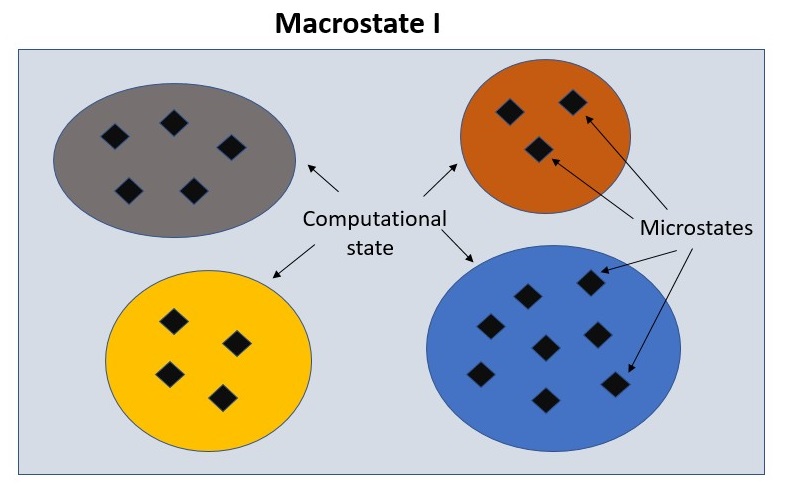}
\end{center}
\caption{A physical system $\mathcal{S}$ with macro-observable/state $I$ is defined by a distribution of microstates. These microstates can then be coarse-grained into computational states (at some intermediate mesoscopic scale) by choosing an appropriate information bearing observable. But we can see how by changing that information bearing variable (and thus how the circles are drawn), we can also have a different set of computational states by coarse-graining over the same microstates.}
\end{figure}

\begin{equation}\label{IB_method}
      \max_{p(k^{\mathcal{S}}|i^{\mathcal{H}})} \mathcal{I}(\mathcal{S}:\mathcal{F}) - \lambda \mathcal{I}(\mathcal{S}:\mathcal{H})
\end{equation}

\noindent where $\mathcal{I}(\mathcal{S}:\mathcal{F})$ is the mutual information between the current system $\mathcal{S}$ and future inputs $\mathcal{F}$ (which we assume is temporally correlated to $\mathcal{H}$), and serves as a measure to quantify the ability of the system to predict $\mathcal{F}$. The trade-off parameter $\lambda$ acts as pseudo-temperature noise parameter (that might be related to the actual temperature $T$). This is the (past-future) information bottleneck problem and it analyses the optimal trade-off between prediction and memory. The solution for the constrained optimization is understood and can be found in \cite{Still3}, \cite{Tishby} and \cite{Tishby2} (Note that if we add to the system $\mathcal{S}$ the ability to act on it's environment and manipulate the input distributions the system will see, then we can obtain state transition probabilities in the system dynamics that exhibit exploitation-exploration trade-offs \cite{Still4}). The main takeaway is that the optimal encoding of the system will include those parts of the past inputs that is relevant to best predicting future inputs and minimize discard irrelevant information. The formulation only shows the existence of computational states in the system that can achieve prediction and not how to build them. What this macrostate could be is hard to identify immediately, but we can say that macro-observable that is being maintained homeostatically cannot be the informationally relevant variable which is used to partition the space of microstates and identify the computational states.

The above equations draw a direct line between a specific NE thermodynamic description of an homeostatic system to an optimally predictive encoding of information in the system states. This brings together keys concepts of \textit{non-equilibrium fluctuation theorems, homeostasis, energy dissipation, optimal encoding and prediction} in a single condition. It is also important to note that while a system $\mathcal{S}$ satisfying the thermodynamic condition of low $\phi_{I \rightarrow I}=\psi_{I \rightarrow I}$ will satisfy the solution for the information bottleneck problem, the reverse is not necessarily true. Systems that implement the information bottleneck algorithm need not satisfy the thermodynamic condition, as evidenced by running these algorithms on existing computing hardware. Thus satisfying low $\phi_{I \rightarrow I}=\psi_{I \rightarrow I}$ is a not a functional condition but a physical thermodynamic one. Also Eq.(\ref{Main}) is a more general equation of homeostasis, whereas Eq.(\ref{IB_method}) corresponds to a very special case of that. Thus we can have a very wide range of homeostatic systems that do not show optimal predictive inference like brains do. With these thermodynamic descriptions for memory and prediction in place, we are now in a position to prescribe the non-equilibrium descriptions of consciousness in physical systems.

\section{Non-Equilibrium Thermodynamic Framework for Consciousness}
In sections 2 and 3, we examined the philosophical underpinnings of the current (popular) frameworks of consciousness, teased out the advantages and challenges of these frameworks and then described a set of criterion (N1-N4) that we would like for useful description of consciousness to have. Before we present the non-equilibrium thermodynamic conditions for consciousness, let us address some of the earlier ideas in this space and how the current work is different from those. The authors in \cite{Solms1}, \cite{Solms2} study consciousness using the popular free-energy principle (FEP) \cite{Friston1}, which models systems using an information-theoretic free-energy term that is minimized. It is important to understand that this is an information theoretic free-energy and not the thermodynamic free-energy, and it is a functionalist framework. Also the authors state in \cite{Solms1} -\emph{`` A fundamental property of living things (i.e., biological self-organizing systems) is their tendency to resist the second law of thermodynamics.''} This is a confusing statement since the ability of biological self-organizing systems to maintain themselves is in accordance with/driven forward by the second law of thermodynamics (and there is no resisting it). Such statements on top of the free-energy minimization formulation makes the author wonder if the authors are using an older (inequality) versions of the second law (Ideas in non-equilibrium thermodynamics are relatively newer and can be a little siloed). Viewing the second law as a continuous entropy increase, a slow march of the thermodynamic arrow of time towards equilibrium (and corresponding free-energy minimization) is valid only when applied to isolated (and some closed) systems. Equilibrium thermodynamics is simply not the right way to study open biological systems that constantly exchange matter and energy with their environments and maintain a non-equilibrium steady state. These systems are better modeled using non-equilibrium fluctuation theorems which allow for a much richer range of behavior and for which no universal extremal principle (like free-energy minimization or entropy maximisation or minimal dissipation) exist as of yet. Other work in this space such as \cite{Yufik}, \cite{Pepperell} both identify the importance of energy flow in the brain, energy efficient and optimal encoding but do not make concrete connections between the thermodynamic relations and the information processing frameworks. It is necessary to mention that the importance of homeostasis in the brain and to consciousness have also been explored in detail in \cite{Damasio} (and we would be remiss not to mention the importance of the good regulator theorem \cite{Conant} and it's role in cybernetics \cite{Ashby}), but we have a stronger physically grounded mathematical formulation that goes beyond existing work by focussing on the non-equilibrium fluctuation theorems.

We will next present the main hypothesis in this paper - two \textit{non-equilibrium thermodynamic conditions of consciousness (TCC)} derived from NE fluctuation theorems, and ties it directly to the information processing in the system. The conditions for a system $\mathcal{S}$ to be conscious is (we can also see this as a condition a system needs to satisfy to be capable of consciousness i.e. proto-consciousness) -   

\begin{itemize}
    \item [(TCC1)] System $\mathcal{S}$ should exhibit dissipation-driven adaptation i.e. forming of correlations between the system state and the external environment, and this produces a finite amount of memory $\mathcal{I}(\mathcal{S}: \mathcal{H})$.
    
    \item[(TCC2)] In addition to TCC1, the system should also exhibit the NE thermodynamic condition of low $\phi_{I \rightarrow I}=\psi_{I \rightarrow I}$ with respect to an homeostatic macro-observable $I$ over a time interval of interest $\tau$ (thus utilizing the memory $\mathcal{I}(\mathcal{S}: \mathcal{H})$). 
\end{itemize}

Both conditions above are not a description of the system in on itself. They are instead description of dynamic processes that a system should be capable of exhibiting that the author proposes as necessary for consciousness. As mentioned in the previous section, satisfying TCC1 corresponds to a system being able to achieve (and maintain) certain degree of complexity/memory in the system. But this condition alone is very broad as it simply requires a system to be able to form correlations with the driving signals from the environment and be capable of having some information content. A lookup table that can be continuously updated with new information will satisfy this condition as it `learns/adapts' to the new inputs and possesses a finite amount of memory of the past inputs. A burnt piece of wood (with say some degree of complexity) has some `memory' of the fire that burnt it, as does a mountain of the river that erodes it in it's current structure. Of course, we do not expect that a lookup table, a burnt piece of wood or an eroded mountain as conscious since none of them are able to exploit the information that they possess in their memory to maintain themselves. This is where TCC2 comes in and thus represents the more important (distinguishing) condition. By requiring a system to be able to satisfy TCC2 over some spatial and temporal scale (in addition to TCC1) constrains the systems that we would identify as conscious. Thus TCC2 is what makes the thermodynamic description a step forward from existing frameworks, and we will focus on it in greater detail in the next section.

\section{Observations on TCC2}
In this section we present some important observations and implications of TCC2 and the larger thermodynamic framework -

\begin{itemize}
  
\item[(a)] The immediate implication of a system satisfying TCC2 is that the system will exploit the information in the memory of the system in order to predict the future (and act on it if the system has the capability) by encoding only those parts of the past information that is relevant. This ability to perform temporal predictive inference combined with the (dissipation driven) memory from TCCI will enable the system to exhibit the type of input-output behavior we would associate with conscious systems. As noted in the previous section, a system satisfying TCC1 and TCC2 will exhibit the necessary input-output relations (of a certain level of intelligent behavior (A1)) but the opposite need not be true. Not all physical systems that exhibit the same functions necessarily satisfy TCC2. In fact we would expect modern machine learning systems that exhibit a certain level of intelligence but are extremely computationally and energetically expensive to not satisfy TCC2. Unlike functionalism, TCC1 and TCC2 being thermodynamic conditions are not implementation/substrate independent and thus clearly differentiate between systems implementing the same functions. If the system also has the ability to act on it's environment, then satisfying TCC2 will imply that the system will exhibit exploitation-exploration dynamics \cite{Still4}, \cite{Clark1} as a part of the optimal solution, and satisfy criterion (A2). 

\item[(b)] Building off the previous point, we would ask - what does TCC2 imply for the structure of the physical systems that realizes the condition? Does TCC2 unconditionally associate consciousness with recurrent or feedforward network structures? The short answer is - \emph{it depends}. This aspect of TCC2 is particularly interesting to explore. Clearly TCC1 simply quantifies the memory capacity of the system and does not limit the structure. While TCC2 does not directly specify one type of structure over another, it is reasonable to assume that not all structures of all substrates can realize the necessary TCC2 condition in a particular type of environment. Thus it might be the case that organic carbon-based biomolecules can only satisfy TCC2 in recurrent neural network structures under our existing environmental conditions. However one cannot rule out the possibility that a feedforward neural network manufactured out of superconducting material from satisfying both TCC1 and TCC2 at near absolute zero temperatures (a likely scenario given the interest in building superconducting neuromorphic circuits for machine learning applications) could be \emph{conscious}. However it would explain why we do not find conscious systems with feedforward \emph{brains} made out of low temperature superconducting material in our regular room temperature environments that we exist in. Thus these thermodynamic conditions correspond to the idea of \emph{constrained functionalism} that was described earlier in this paper. We could also build falsifiability tests around this idea - in addition to experimentally validating TCC2 in conscious systems (like humans and some animals), if we can engineer feedforward networks that satisfy TCC2 using organic bio-molecules under acceptable environmental conditions at relevant spatial scales that are indistinguishable from humans, that would raise serious questions on why feedforward conscious human brains did not naturally evolve and/or serve as a way to falsify TCC2 as a condition for consciousness. 

\item[(c)] This aspect of TCC2 on acceptable network architectures makes it distinct from IIT which would have ruled out consciousness for any system with a feedforward architecture. Interestingly when faced with the possibility that IIT might not be falsifiable given it's inability to validate the value of$\phi$ between equivalent recurrent and feedforward neural networks \cite{Doerig}, some advocates of IIT invoked an efficiency-based argument that feedforward systems would be much less efficient than their corresponding recurrent equivalent and this hence favors IIT's claims on recurrent networks (even this does not quite work out given the results in \cite{Walker}). However IIT as currently formulated has no way to justify this efficiency argument other than simply positing it for convenience on top of their existing axioms and postulates. On the other hand the minimal dissipation associated with TCC2 is an inherent energy efficiency and optimal representation condition. Thus under the thermodynamic descriptions, it is reasonable to compare and distinguish between both the efficiency and conscious status of a feedforward vs recurrent network implemented using the same substrate. In fact, we should also be looking to further explore the relationship between certain network topologies in the human brain, their relationship to information transmission and compression efficiency, and metabolic costs in the brain \cite{Bassett1}, \cite{Bassett2}.

\item[(d)] TCC2 is a coarse-grained macroscopic description allowing for multi-realizability through a large number of trajectories of microscopic states at different spatial and temporal scales. It is also an emergent description in the sense that `consciousness' could be better described at the macroscopic thermodynamic condition than at the level of the underlying microscopic trajectories. Another important advantage of TCC2 is that it provides a condition for consciousness as a dynamic process with an associated spatial and temporal scale, as opposed to simply a state of the system. The condition in question - low $\phi_{I \rightarrow I}=\psi_{I \rightarrow I}$ has a spatial resolution through the choice of the macroscopic variable $I$ and a temporal resolution through the choice of the time interval $\tau$ of interest to us to study. Through the mapping of TCC2 to the information bottleneck problem, we can see the importance of input signals that the homeostatic system is resisting changes to and is thus predicting - these become the signals the system is `conscious' of. The question isn't anymore of asking if certain states of a system are conscious, sub-conscious or unconscious. It is instead one of asking if the dynamical trajectories of states in a system at certain spatial and temporal resolution of interest under specific inputs satisfy the necessary condition. Given the spatial and temporal dependence, the thermodynamic framework also does not describe consciousness as an `all-or-nothing' phenomenon. It is a more gradual phenomenon within a system depending on a number of parameters (for eg: the trade-off pseudo-temperature parameter $\lambda$ from Eq.(\ref{IB_method}) whose effect on optimal-coding of input signals has been studied in \cite{Still3} and discussed below in (f)), as well as across a wide range of systems. 

\item[(e)] The mapping of TCC2 to the information bottleneck problem allows us to combine it with some powerful existing frameworks in neuroscience that has been used to understand various aspects of cognition, perception, emotion, etc. FEP is one such popular framework used to model various cognitive phenomenon at multiple spatial and temporal scales \cite{Friston1}, \cite{Friston2}, and it's relationship to the information bottleneck problem is established in \cite{Friston3}. It would require a good amount of work but it seems very plausible to map the equations above to the predictive coding/processing (PP) frameworks \cite{Rao}, \cite{Clark}, \cite{Seth1}, \cite{Seth2}. This would offer another powerful tool in neuroscience and the wealth of knowledge that they have already accommodated, that we should be looking to explore and exploit. Also of great interest to study would be the information generation hypothesis from \cite{Kanai2}, which views the function of consciousness as generating counter-factual information. Now if you put that along side temporal prediction and ability to act (inervene) under exploitation-exploration dynamics for an active agent, we get Judea Pearl's causal ladder, which he considers as necessary for human-level intelligence \cite{Pearl}. The mapping of TCC2 to these functionalist models seems slightly easier than mapping these functionalist models to the brain neurobiology. Both of these projects will involve a lot of work across multiple disciplines.

\item[(f)] In the solution for the encoding probabilities, the $\lambda$ noise parameter from the Eq.(\ref{IB_method}) is like a pseudo-temperature (probably dependent on the actual temperature, noise in the system, etc) and gives a range of encoding strategies as it is varied from very low (deterministic/large order) to very high (random/no order), not all of which are good for information encoding. The variation of such pseudo-temperature parameter and their effect on the encoding strategies used by spiking neurons has been studied in \cite{Still3} and \cite{Gasper}. We can also view the variation in the encoding in the system as $\lambda$ is varied as moving the system between sub-critical and super-critical phases. Study of the information bottleneck method has shown that solutions exhibit a number of critical points with respect to $\lambda$ and merits greater work to understand it's relationship (if any) to the self-organized criticality hypothesis in the brain \cite{Bak}, \cite{Hesse}. The different states of consciousness can be thus be viewed as maintaining TCC2 but having $\lambda$ manipulated, hence affecting how information is stored in the system and used to achieve prediction. Viewing it as changes to the spatial and temporal scale in which TCC2 is achieved adds an extra layer of complexity that warrants further work. The changes in the spatial scale could be used as explain how information could be stored/represented in the brain but not be part of our conscious awareness. Assume that there is a no (significant) spatial overlap in the part of system $\mathcal{S}_0$ which satisfies TCC1 and stores a piece of information, and the part $\mathcal{S}_1$ that satisfies both TCC1 and TCC2 and thus makes predictions using the finite memory in $\mathcal{S}_1$. Thus $\mathcal{S}_1$ cannot exploit the information available in $\mathcal{S}_0$ to make predictions, and thus the conscious experience will lack that information even though it has been stored in the system. However if the joint system $\mathcal{S}_0\mathcal{S}_1$ is satisfying TCC2, we would expect the incoming information in both systems to be part of our conscious experience. The author wants to reiterate that an unavoidable consequence of using these thermodynamic conditions will be this high level abstraction stripping away many details.

\item[(g)] TCC2 being a macroscopic thermodynamic condition takes a different view over questions that involve replacing parts of your biological neurons with functionally equivalent silicon-based substitutes i.e. ideas of absent, fading and dancing qualia \cite{Chalmers0}. The system $\mathcal{S}$ is conscious as long as it can maintain TCC2 at a certain spatial and temporal scale with respect to a macrovariable. Thus if $\mathcal{S}$ continues to satisfy TCC2 at a same/different spatial scale even after one of the biological parts has been replaced with functionally equivalent part made of silicon, we would expect the system to be conscious but with a slightly similar/different experience respectively. As the number of silicon neurons are increased, and if the system remain conscious by satisfying TCC2 but at a different spatial scale (and perhaps observable) as compared to a human with only biological parts, then we would expect the system to have different experiences. The author would like to note that this indicates very early thinking on this question and more work is needed to address changes in the contents of experience.

\item[(h)] Satisfying TCC2 simply states that there exists a partitioning/coarse-graining of the system microstate (trajectories) into computational states that satisfy Eq.(\ref{IB_method}). The condition does not specify how to construct this partition and thus makes no assumption on what components of the system could constitute the relevant computational states. For eg: Traditional ideas tend to focus on neurons and synapses in the human brain to build computational models. However it could be the case that synapses, dendrites, neurotransmitters, proteins, glial cells, etc. can all be involved in achieving TCC2 and will require detailed experimental work in the construction of these relevant states.

\item[(i)] Philosophers often talk about the need for frameworks of consciousness to discuss our experience of time. The author not have any experience in this specific topic (and is not the focus of this work) and cannot offer any detailed comments. However we will point out that the very arrow of time is a thermodynamic description and by placing consciousness in the same language as time, we might have a way to addressing it rigorously and in accordance with physical law.

\end{itemize}

\section{Stance on the Hard Problem}
No framework that looks to address consciousness can be considered complete without offering a stance on \emph{the hard problem of consciousness} introduced by Chalmers \cite{Chalmers1}, \cite{Chalmers2}. The hard problem of consciousness is the problem of explaining \emph{qualia} or phenomenal experience - `why does it feel like anything when we are conscious.' Chalmers differentiates the hard problem from the `easy' problems of consciousness which includes perception, attention, ability to discriminate, etc. It is a reasonable to question why TCC1 and TCC2 are not simply conditions that addresses the easy problems, and nothing more. The author believes that this is not the case here. The history of artificial intelligence, especially the recent successes in machine learning (ML) through the use of artificial neural networks have shown us that it is possible to build systems that can solve narrow \emph{intelligence} tasks like Go, driving (to a certain extent), face recognition, etc with clever computing ideas and availability of resources. No one in the community seriously thinks these systems are in any manner conscious. As pointed out before, a system that can implement the necessary functions to mimic an intelligent system does not immediately satisfy the thermodynamic conditions of consciousness prescribed above (but the opposite is true based on the TCC1 dependent memory constraints). The exponential (super-Moore level of) growing demand of compute power and associated energy costs is a serious problem for the ML community. The human brain (a system that we agree is conscious) is often pointed to as the holy grail achievement for the AI field - given it's ability to be intelligent across tasks and efficiency of it's implementation. Since we could in principle achieve performance across tasks given enough data and compute resource, one has to ask if energy efficiency that the human brain exhibits plays a crucial role here. Now TCC2 brings together task-independent temporal prediction and energy efficiency together in a single condition. All we can look to do is provide a description that brings these two key properties of an agreed-upon conscious system together (one of which we know we can achieve in principle without the energy efficiency constraint) and ask if that description captures the necessary phenomenon.

The intuition that these conditions simply seems to only solve the easy problems is not misplaced. As it stands now, there still seems to be a huge gap between what the thermodynamic framework is discussing and the \emph{phenomenal properties} that form an important part of our experiences. How are these conditions possibly a solution the hard problem? With respect to the hard problem, the thermodynamic framework takes an \emph{illusionist} position \cite{Dennett},\cite{Frankish1}, \cite{Frankish2}. The illusionist stance is to reject the framing of the hard problem outright, given the various issues associated with the realist position towards solving the hard problem. Illusionism states that consciousness is real and the subjective qualities you experience are also real, but they are not what you think they are. It rejects the existence of non-physical mental states or that of phenomenal properties (distinct from physical properties). What is illusory is our introspective experience of these properties as being non-physical and different from what we perceive of our outside world. To the illusionist position, the meta-problem of consciousness \cite{Chalmers3} is the problem of consciousness \cite{Frankish3} and thus solving the meta-problem dissolves the hard problem. Before we proceed further, I want to quote a few lines from \cite{Frankish1} that I think are extremely important and clarifies the illusionist position (which given the use of useful plus provocative term illusion has often been easy to misunderstand and straw-man) - \emph{``The third option is illusionism. This shares radical realism’s emphasis on the anomalousness of phenomenal consciousness and conservative realism’s rejection of radical theoretical innovation. It reconciles these commitments by treating phenomenal properties as illusory. Illusionists deny that experiences have phenomenal properties and focus on explaining why they seem to have them. They typically allow that we are introspectively aware of our sensory states but argue that this awareness is partial and distorted, leading us to misrepresent the states as having phenomenal properties. Of course, it is essential to this approach that the posited introspective representations are not themselves phenomenally conscious ones. It would be self-defeating to explain illusory phenomenal properties of experience in terms of real phenomenal properties of introspective states. Illusionists may hold that introspection issues directly in dispositions to make phenomenal judgments — judgments about the phenomenal character of particular experiences and about phenomenal consciousness in general. Or they may hold that introspection generates intermediate representations of sensory states, perhaps of a quasi-perceptual kind, which ground our phenomenal judgments. Whatever the details, they must explain the content of the relevant states in broadly functional terms, and the challenge is to provide an account that explains how real and vivid phenomenal consciousness seems.''} In particular the author will advocate for strong illusionism as defined by Frankish \cite{Frankish1} \footnote{It has been brought to the author's attention that there are different taxonomies in illusionism - Frankish's \cite{Frankish1} vs Chalmers \cite{Chalmers3}. To clear any doubt the author will refer to Chalmer's version of strong illusionism which states that `consciousness is really an illusion and it doesn't exist' as \emph{stark illusionism}, and when we refer to strong illusionism, it is strictly how Frankish defines it.}, that introduces the idea of quasi-phenomenal properties \cite{Frankish1} - \emph{``A quasi-phenomenal property is a non-phenomenal, physical property (perhaps a complex, gerrymandered one) that introspection typically misrepresents as phenomenal. For example, quasi-phenomenal redness is the physical property that typically triggers introspective representations of phenomenal redness.  There is nothing phenomenal about such properties — nothing `feely' or qualitative — and they present no special explanatory problem. Strong illusionists hold that the introspectable properties of experience are merely quasi-phenomenal ones.''}

In order to understand how the thermodynamic framework reaches the illusionist stance, let us go back and review TCC2 again. From section (4.2), we have that the special case of low $\phi_{I \rightarrow I}=\psi_{I \rightarrow I}$ is equivalent to an information bottleneck in Eq.(\ref{IB_method}).

\begin{center}
      $\max_{p(k^{\mathcal{H}}|i^{\mathcal{S}})} \mathcal{I}(\mathcal{S}:\mathcal{F}) - \lambda \mathcal{I}(\mathcal{S}:\mathcal{H})$
\end{center}

Illusionism has pointed to the lack of sufficient resources in the brain as the reason for the emergence of quasi-phenomenal states \cite{Frankish2}. We will adopt a similar approach here. The information-bottleneck can be really seen as a compression of the information \cite{Tishby} to form coarse-grained representations \cite{Wolpert} under memory constraints imposed through TCC1 and energy efficiency constraints imposed through TCC2. The identification of macro-observable $I$ that satisfies TCC2 determines the input history $\mathcal{H}$ and future inputs $\mathcal{F}$, as well as the spatial scale of $\mathcal{S}$ of the conscious system. This decides what would be inputs to the system (thus input history) and the memory capability of the system $\mathcal{I}(\mathcal{S}:\mathcal{H})$. As the system generates coarse-grained representations (compression) of the information in the inputs to $\mathcal{S}$ in order to perform predictive inference of future inputs, different parts of the inputs are going to be compressed to different degrees. There is a gradation in the scale of coarse-grained representations generated in $\mathcal{S}$. 

We will explore this a little bit further here. Chalmer's demarcation makes a strong distinction between those problems he considered easy (E) and those which were hard (H), requiring a radical shift on the ability to science to address problems on either side of that boundary. On the other hand, the author is making the case that consciousness - both access and phenomenal is very real but instead of clear boundary placing each of those in discrete categories, we propose a more gradual scale with those parts of consciousness that we regard as part of the easy problems at one end and those we regard as hard on the other, and different parts of our experience spread across the scale. Everything on the scale is a physical process and there are no physical/mental distinctions. To make the analogy to human experience about this scale - certain features of our external sensory perception on one end of the coarse-graining to highly coarse-grained \textit{interoceptive-plus} (from the body plus other parts of the brain outside of the $\mathcal{S}$ satisfying TCC2) inputs that add the subjective `feelness' to our experiences. There is no special ontological status to the more compressed signals and it is as real and physical as other parts of the sensory perception. It has a mysterious \textit{subjective quality} to it, due to severe coarse-grained compression to the point that it becomes unrecognizable and distinct from the physical, and we mistake them to be mental/phenomenal and invoke ideas of \emph{qualia}. When we pay attention to these through introspection, these inputs are so heavily coarse-grained in our system as compared to certain parts of the sensory signals that any verbal report (on say describing the 'feeling' of red) will not have the same amount of clarity as describing other parts of our perception (It is in fact hard to decouple the mysterious subjective feeling in experiences without seemingly less subjective i.e. the 'physical' aspects providing a reference). The generation of body ownership and of the concept of self or `I' under a predictive processing framework \cite{Seth2} are coarse-grained representations as well, that lie somewhere along this scale between the two extremes. A closely related idea of Bayesian frugality \cite{Dewhurst2} has also been proposed to bring together predictive processing framework and the Attention Schema Theory \cite{Graziano} to explain the \emph{mysterious subjective} aspects of experience. All of these culminate together to form our \emph{first person view} as the inevitable result of imposing resource constraints (realized as information bottlenecks) on a constraint-free \textit{third person view}. The third person description of a system realizing these conditions is clearly not the same as the system actually realizing the conditions (and thus having the experience) and indicates the explanatory gap \cite{Levine} that seems to exist between the two.  

Building off the previous idea, how should one think about the content of consciousness under the thermodynamic framework?  First TCC1 and 2 should be seen as grounding conditions to establish whether or not a system is conscious. Once we do that, the conscious content itself (atleast in humans) is most usefully described by existing functional descriptions. While the thermodynamic framework is not a functionalist framework, the claim here is that - \emph{experiential states (trajectories) are best described functionally in systems that we agree to be conscious}. We should be looking for families of trajectories of physical microstates over time and work to establish relationships between inputs, these state trajectories and the functional labels (obtained through behavioral reports) that we traditionally use to identify these states \footnote{We will leave the discussion around contents of experiences of artificial systems for a different work}. In these suitably identified partition of microstates into computational states, we have coarse-grained representations (to varying degree) of the different (interoceptive, proprioceptive, exteroceptive, etc) inputs to the system that is used to generate predictions (of the same) forming the content of our conscious experience. As discussed in the previous section, there has been tremendous work in these areas using functionalist frameworks like predictive coding, and the next important step is formalize the relationship between the thermodynamic and functionalist pictures with more rigor. 

\section{Some Major Challenges That Lie Ahead}
Having presented an highly optimistic and favorable view of the thermodynamic framework, the author thinks it is only fair to point a couple of (major) challenges and problems that are immediately identifiable. The author hopes that by making a positive case for a new thermodynamic framework of consciousness, it will encourage a number of more intelligent researchers to explore this in further detail and address some of these issues - 

\begin{itemize}
    \item [(a)] While the author has assumed the existence of macroscopic $I$ that (should) satisfy TCC2 in conscious systems, we do not know of a procedure to identify them, especially in complex biological systems like humans. We might look to use literature on metabolic studies in the brain to perhaps identify potential candidates? Furthermore once we identify the observable $I$, there is still the question of grouping these microstates into computational states, an unenviable task given the complexity of known conscious systems. Furthermore  anyone working in thermodynamics will immediately attest to the difficulty of running simulations of non-equilibrium systems. Trying to map trajectories is computationally very hard and possibly intractable for any mesoscale to macroscale system of interest. While a thermodynamic framework of consciousness might be elegant and potentially useful given it's relation to physical law, the downside might be that by abstracting to the level of thermodynamic descriptions to achieve generality across systems, we might lose the ability to exploit existing computational tools as well as map them onto biological systems to answer questions that neuroscience is interested in. But as was the case with IIT, one would hope that people will come up with clever ways to find proxies for TCC2 or identify ways to use techniques like Markov chain Monte Carlo to make the problem tractable. The goal of these frameworks are never meant to replace the wealth of knowledge that neuroscience can deliver at computational and neurobiological levels. They are meant to provide a bigger picture view to fit all the pieces together in accordance with physical law (and make a strong case against adoption of a position that consciousness lies outside the realm of science). 
    %cite metabolic studies here
    
    \item [(b)] This second (and more pressing) problem is related to the \emph{'scale problem of consciousness'} - what is the level at which consciousness is realized \cite{Kanai}? Let us assume that we are able to identify $I$ in a system $\mathcal{S}$ that satisfies TCC2 over some time interval $\tau$, and as per the framework proposed in this paper, we classify the system as conscious. Now let there be a subsystem $\mathcal{S}_0$ of $\mathcal{S}$ that has an observable $I_0$ (which of course would be part of the higher level macro-observable $I$ of $\mathcal{S}$) that also satisfies TCC2 at it's level of description for the same time interval. The question now is - what is the level at which there is conscious experience? At the level of $\mathcal{S}$ or $\mathcal{S}_0$ or both? IIT deals with this problem using their \emph{exclusion axiom} and postulate that - \emph{of all overlapping sets of elements, only one set with maximal integrated information can be conscious} \cite{Hoel}. The author can either adopt the same exclusion axiom, but that would not come immediately from physical law like the conditions themselves did. One could take the stance that there would never be a situation in which $I_0$ satisfies TCC2 in $\mathcal{S}_0$ when it is also satisfied by $I$ in $\mathcal{S}$ as well, but this feels like wishful thinking (and the author does not see a reason why this would be impossible in principle). This leaves us with one remaining option - to bite the bullet and take the view similar to the authors in \cite{Kanai} and ``allow for multiple consciousnesses to coexist across different levels of coarse-graining within a system if they are informationally closed from each other.'' Since $\mathcal{S}_0$ is a subsystem of $\mathcal{S}$ as it satisfies TCC2, signals originating from it cannot be inputs outside of $\mathcal{S}$, that the system can be conscious of. Thus even if $\mathcal{S}_0$ does satisfy TCC2 simultaneously, the conscious experience of the larger $\mathcal{S}$ will not include identification of micro-conscious $\mathcal{S}_0$ (or in the very least so extremely coarse-grained that it does not recognize it). By not positing an exclusion principle like IIT 3.0 \cite{Oizumi}, we escape from the some of problems that IIT faces as pointed out in \cite{Cerullo}. But we will use the example used in \cite{Schwitzgebel} to further illustrate our point and make the difference with IIT. The original problem in IIT was the broad allocation of consciousness to the point where a properly arranged networks of all the humans in the USA would create a consciousness of the USA. An anti-nesting principle was then subsequently proposed to solve this problem, according to which when subsystems with non-zero $\phi$ join to form a system with a larger value of $\phi$, the subsystems lose their consciousness. This then led to other problems (counter-intuitive predictions) of its own - such that if a small brain implant was placed in a human brain, integrated in such a manner so that the consciousness of the human+implant is greater than the human, then that would imply that the human would lose their consciousness and be replaced with a larger human+implant consciousness. Instead under the thermodynamic framework, assume that the human ($\mathcal{S}_0$) is able to satisfy TCC2 after integration simultaneously as the human+implant system ($\mathcal{S}$), then $\mathcal{S}$ is conscious as a joint human+implant system of inputs outside of it and not aware of $\mathcal{S}_0$ (which forms a part of $\mathcal{S}$) as a separate conscious entity outside of itself. At the same time, the human $\mathcal{S}_0$ is conscious of itself, experiences the implant as external to it and does not recognize that the joint human+implant $\mathcal{S}$ system is also conscious. 
\end{itemize}    
   
Of course these are not the only possible problems and challenges facing this framework. When dealing with one of the toughest problems in science, it is to be expected that many more issues will pop up. The question is whether or not there is enough interest among talented scientists and philosophers, as well as the availability of resources to solve them.

\section{Machine Consciousness}
One of the most interesting questions in the science of consciousness is the possibility of building machines capable of consciousness? Many philosophers, scientists and engineers have wondered about this very questions for a very long time but it has become increasingly pertinent with the recent success in the field of artificial intelligence and machine learning, with different necessary conditions prescribed \cite{Lau}. As these artificial systems continue to get better at tasks that have been historically associated with our intelligence, it is natural to question the conscious status of these systems \cite{Reggia}? If these systems indeed are conscious, then it represents a monumental achievement for human civilization, as well open a Pandora's box of ethical and legal questions to decide. We will utilize the taxonomy provided by Gamez in \cite{Gamez1} with respect to the levels of machine consciousness in our discussion here -

\begin{itemize}
    \item{\textit{(MC1)}} \textit{Machines with the external behaviour associated with consciousness.}
    \item{\textit{(MC2)}} \textit{Machines with the cognitive characteristics associated with consciousness.}
    \item{\textit{(MC3)}} \textit{Machines with an architecture that is claimed to be a cause or correlate of human consciousness.}
    \item{\textit{(MC4)}} \textit{Phenomenally conscious machines.}
\end{itemize}

Over the years we have made steady progress in constructing models for MC1 and MC2 and realizing them in artificial systems. MC3 is possible in principle and with the increased interest in neuromorphic computing, we should expect to be able to achieve it in the near future if we wanted to. But we have made no progress towards MC4, though a number of theories (like the ones we have discussed earlier) have been proposed. A detailed discussion of theoretical frameworks for human and machine consciousness, their advantages and drawbacks as well what a theory of consciousness should look like is discussed in \cite{Gamez2}. Without a solid framework of consciousness in humans and animals, MC4 has looked near impossible and discussion around it has mostly been dismissed as fiction. Of course popular media depictions of artificial intelligence, robots and artificial consciousness as well as the pseudo-scientific claims about Singularity and consciousness uploading have served to only further damage this area of research.

We will specifically address MC4 in this section (since the other 3 are functional realizations and can be implemented in principle). From \cite{Gamez3}, we have the following four pre-conditions for a generalizable theory of consciousness that would be needed for MC4 -

\begin{itemize}
\item[(1)] Can generate testable predictions.
\item[(2)] Is applicable to any physical system.
\item[(3)] Compact (Occam’s razor).
\item[(4)] Based on objective properties of the physical world.
\end{itemize}

\par\noindent The thermodynamic conditions TCC1 and TCC2 are applicable to any physical system, can be used to generate testable predictions around energy absorbed and dissipated that could be studied experimentally (fMRI, metabolic studies, etc) and in simulation. While the author cannot comment on the simplicity of the conditions, we should note that the thermodynamic conditions are derived from the fluctuation theorems and thus a result of objective physical law and the objects of study would be measures like energy, work, heat as opposed to information and computation. The author proposes that an artificial system satisfying TCC1 and TCC2 can be MC4 conscious in principle. What about a computer that simply simulates the computational equation Eq.(\ref{IB_method})? As we have stated before, systems simulating/implementing Eq.(\ref{IB_method}) need not satisfy low $\phi_{I \rightarrow I}=\psi_{I \rightarrow I}$ since that relationship is not reciprocal and thus need not be conscious. A lot of the current large-scale computational brain models run on a significant amount of compute resources \cite{HBP} almost definitely do not satisfy Eq.(\ref{Main}) (The author has not experimentally verified this), and cannot be expected to magically become conscious as more and more compute is added on.

\textit{What can we say about our current AI systems in von-Neumann vs neuromorphic hardware?} We have to start replacing these type of questions with something more specific, since we are not really asking whether or not AI systems with von-Neumann or neuromorphic hardware made out of any substrate would be conscious or not, under all possible conditions? We are more interested in whether or not a particular systems that we have built or intend to build with specific materials (usually silicon CMOS) operating in our regular environments are conscious? We will have to experimentally check if our AI system $\mathcal{S}_{AI}$ (in both von-Neumann and neuromorphic) satisfies TCC2, as they both almost definitely satisfy TCC1. But my educated speculation is that AI systems built on traditional CMOS devices and implemented using the von-Neumann architecture will definitely not satisfy TCC2 and hence not be conscious for the following reason - if it was possible for such a system to be conscious, one would have reasonably expected to see some evidence of bio-molecular von-Neumann brains somewhere in the tree of life, achieving the same thing as our traditional brains to have evolved as the product of evolution. Questions along this line of inquiry can now be translated from a philosophical thought experiment to the form of research in network theory and material chemistry - \textit{can a carbon-based substrate satisfy TCC1 and TCC2 in any other network structure other than the small-world recurrent one in our heads?}  

Neuromorphic computing represents a wide range of hardware systems built to mimic the brain at various levels of devices, architectures, etc \cite{Schuman}. It is a much harder case as it represents a grey area between CMOS-based von-Neumann architectures on one end and biological brains on the other. It is probably best handled on a case-by-case basis as the answer will depend on multiple factors in the computing stack, as well as on the spatial and temporal scale of interest. On the other hand, I am very interested in self-organizing neuromorphic atomic switch networks (ASN) \cite{Steig} - it represents a very interesting candidate to test these ideas for artificial consciousness, since they represent a lot of the right properties you would expect to see in a system that satisfies TCC 1 and 2 like - memristive device behavior (memory), self-organizing scale-free network architecture and can be used perform time-series predictions in an energy efficient manner. There is increasing interest in material science research to fabricate more novel materials, similar to the silver nanowires in the ASNs to produce self-organizing networks for AI tasks. The author believes that this is a very interesting overlap between those interested in machine consciousness and those building these novel energy efficient AI hardware. Increasing demand for the latter (and the resources that come with that demand) might rejuvenate the earlier.

\section{Summary \& Conclusion}
In this paper, we proposed that a thermodynamic description of consciousness would be a step forward from both functionalist and causal structure frameworks. Then starting from the non-equilibrium thermodynamic fluctuation theorems (physical law), we proposed two conditions for consciousness - TCC1 (dealing with memory) and TCC2 (dealing with efficeint prediction-memory trade-off under homeostasis). The advantages of these conditions over existing frameworks, it's connection to existing functional theories and it's stance on the hard problem is discussed. We finally ended with our predictions on the questions of machine consciousness and the path forward. To the best of my knowledge, this work here represent the first framework of consciousness (both biological and artificial) that is derived straight from existing non-equilibrium thermodynamical physical law (and does not need to appeal to any new physics) and brings together ideas from philosophy, physics, information theory, functional predictive coding models and neuroscience. It seeks to put consciousness squarely at the heart of physical law and in the realm of science, rather than outside it (and does not try to replace other modes of studying human brain function/consciousness like neuroscience and cognitive science, which is in a better position to find solutions to problems that have to do with cognitive disorders, etc). It can however try to answer questions like`why did human consciousness emerge?' by translating them to into the realm of science - \emph{'what is the probability of forming a structure at the necessary spatial scale that satisfies TCC2 over evolutionary time scales?}. I would like to end by reiterating that there is still a lot of work to be done but the future of the scientific study of consciousness looks very bright indeed.

\label{sec:headings}

\bibliographystyle{unsrt}  


\end{document}